# OCLEP+: One-class Anomaly and Intrusion Detection Using Minimal Length of Emerging Patterns[*]

Guozhu Dong and Sai Kiran Pentukar
Department of Computer Science and Engineering
and Kno.e.sis Center
Wright State University, Dayton, Ohio 45435, USA
Email: guozhu.dong at wright.edu

November 21, 2018

**Abstract**

In an earlier paper, a method called One-class Classification using Length statistics of Emerging Patterns (OCLEP) was introduced. This paper presents OCLEP+, which is a modified version of OCLEP, for constructing one-class anomaly detection and intrusion detection systems. OCLEP+ mainly differs from OCLEP in what length statistics is used and how the length statistics is derived. OCLEP+ has these main features: "one-class training," "not using model/signature" (making it hard for adversaries to attack), "not using distance metrics", and "using length statistics of patterns" (making it a robust method). OCLEP+ can be used for one-class intrusion detection, one-class anomaly detection, one-class masquerader detection, and one-class outlier detection. Experiments show that OCELP+ is more accurate than OCLEP and one-class SVM on the NSL-KDD dataset.

## 1 Introduction

Intrusion detection plays an important role for ensuring the security of various protected systems. Intrusion detection is a special case of anomaly detection, and it is also related to novelty detection and outlier detection. Very often the underlying data for intrusion detection have high dimensions and perhaps also contain many categorical attributes. Many existing intrusion detection and anomaly detection approaches directly rely on the use of distance metrics, while some others indirectly rely on distance metrics (e.g. they use models that involve attribute-wise differences between data points). These approaches may perform poorly due to pitfalls of distance metrics in high dimensional spaces (see Section 2) and due to the existence of categorical attributes.

---

[*]Citation: Guozhu Dong and Sai Kiran Pentukar. OCLEP+: One-class intrusion detection using minimal length of emerging patterns. CSE Tech Report, Wright State University, 2018.

In an earlier paper [6], a method called One-class Classification using Length statistics of Emerging Patterns (OCLEP) was introduced.

This paper presents OCLEP+, which is a modified version of OCLEP, for constructing one-class intrusion detection and anomaly detection systems. OCLEP+ mainly differs from OCLEP in what length statistics is used and how it is derived. OCLEP+ retains the main features of OCLEP, including "one-class training", "not model or signature based" (making it hard for adversaries to attack), "using length statistics of patterns" (making it a robust method), and "not using distance metrics." OCLEP+ and OCLEP can both be used for one-class masquerader detection, one-class intrusion detection, one-class anomaly detection, and one-class outlier detection. The patterns identified for given instances predicted to be intruders can be used to explain why predicted intruders are intruders. Experiments show that OCELP+ is more accurate than OCLEP and one-class SVM on the NSL-KDD datasets.

Organizationally, the paper first gives some general background for intrusion detection, followed by some preliminaries. Then it presents the emerging pattern length based method **OCLEP**+ for intrusion detection. Next it reports experimental evaluation of the method on a public dataset which has been widely used to evaluate intrusion detection systems. Finally some concluding remarks are given.

## 2  Background on Intrusion Detection and Anomaly Detection

Since intrusion detection is a special case of anomaly detection (especially when the training data are all from one class), we present some background on both topics.

Anomaly detection is concerned with identifying anomalies with respect to some given dataset. Anomaly detection can be treated as two-class problems and one-class problems. In the two-class case, a dataset containing two classes, representing normal instances and anomaly instances respectively, is given to train the anomaly detection system. For the one-class case, a dataset containing just one class (representing the normal instances) is given to train the system. One-class anomaly detection is also called *outlier detection*. The one-class case for anomaly detection is more challenging than the two-class case. On the other hand, it is easier to use as the training data for one-class is easier to obtain, and detection systems based on one-class training data can be deployed earlier in the life-cycle of the application.

For network and system security, anomaly detection includes intrusion detection and masquerader detection. The former is about detecting intruders to a protected system, while the latter is about identifying abusers inside a protected system who use privileges of others through non-authorized means.

Reference [4] gives a survey of the machine learning and data mining literature on intrusion detection. It indicates that most methods are classification model based methods, although clustering based methods have also been studies. A notable statement from the paper is the following:
"Unfortunately, the methods that are the most effective for cyber applications have not been established; and given the richness and complexity of the methods, it is impossible to make one recommendation for each method, based on the type of attack the system is supposed to detect."

Reference [16] provides a survey on outlier detection. From this survey, one can see that most existing outlier detection approaches rely on the use of distance metrics, while some others rely on models that involve the use of data-point differences. Reference [3] points out certain pitfalls of distance metrics in high dimensional space, asking the question "When is 'nearest neighbor' meaningful?" Reference [15] contains a discussion of some more recent outlier detection algorithms. Reference [1] discusses challenges to outlier detection in high dimensional space.

# 3 Preliminaries on Jumping Emerging Patterns

We first give some relevant basics on patterns. An *item* is a *single-attribute condition* of the form "$A = a$" if $A$ is a categorical attribute, or "$v_1 \leq A < v_2$", where $v_1$ and $v_2$ are constants, if $A$ is a numerical attribute. A *pattern* or *itemset* is a finite set of items. An instance $x$ is said to *satisfy*, or *match*, a pattern $P$, denoted by $x \models P$, if $x$ satisfies every item/condition in $P$.

Given a dataset $D$, the *matching data* of $P$ in $D$ is given by $\mathsf{mds}(P, D) = \{x \in D \mid x \models P\}$. The *support* of $P$ in $D$ is $supp(P, D) = \frac{|\mathsf{mds}(P,D)|}{|D|}$.

In this paper, we use a special kind of jumping emerging patterns [9]. Here, one of the classes contains just one instance.

> Given a data instance $t$ and a dataset $D$, a *jumping emerging pattern* for $t$ is a pattern $P$ that occurs in $t$ but not in any instance in $D$. $P$ is referred to as a JEP for $t$ vs $D$.

To reduce redundancies, we limit ourselves to minimal[1] jumping emerging patterns for $t$ vs $D$.

For example, for $t = 1234$ (representing $\{1,2,3,4\}$) and $D = \{3456, 2479, 2358\}$, there are two minimal JEPs, namely 1 and 234.

---

[1] Minimal in the set containment sense

The discussion below also refers to the general concept of emerging patterns, which we define in the remainder of this section. Let $D_1$ and $D_2$ be two data groups. The *growth rate*[2] [9, 10] of a pattern $X$ from $D_i$ to $D_j$ ($i, j \in \{1, 2\}$ and $i \neq j$) is

$$\mathsf{growthRate}(X, D_j) = \mathsf{supp}(X, D_j) / \mathsf{supp}(X, D_i).$$

For the case $\mathsf{supp}(X, D_i) = 0$, it is customary to

- define $\mathsf{growthRate}(X, D_j)$ to be 0 if $\mathsf{supp}(X, D_j) = \mathsf{supp}(X, D_i) = 0$, and
- define $\mathsf{growthRate}(X, D_j)$ to be $\infty$ if $\mathsf{supp}(X, D_j) > 0$ and $\mathsf{supp}(X, D_i) = 0$.

Given a growth-rate threshold $\rho > 0$, a pattern $X$ is a *$\rho$-emerging pattern* for data group $D_j$ if $\mathsf{growthRate}(X, D_j) \geq \rho$. If $X$ is an emerging pattern for $D_j$, then $D_j$ is the *home data group* (also called *target data group*) and the other dataset is the *opposing data group* (also called *background data group*) of $X$. An emerging pattern having $\mathsf{growthRate}(X, D_j) = \infty$ is called a *jumping emerging pattern* for $D_j$.

## 4 The OCLEP+ Method

The *length* of a pattern is defined as the number of single-attribute conditions contained in the pattern. For example, the length of *ace* (={*a, c, e*}) is 3.

Below we use $N$ to denote a training dataset of normal instances.

### 4.1 An Observation on Emerging Pattern Length

Normally one would use emerging patterns to differentiate instances in different classes. It turns out that one can also use emerging patterns to differentiate instances that are from a common class. One would likely observe the following difference:

> The emerging patterns differentiating instances of different classes are often short, whereas the emerging patterns differentiating instances of a common class are often long. The former are typically much shorter than the latter.

Experiments [6] on the Mushroom data from the UCI Repository confirmed this observation: The average length of minimal jumping emerging patterns differentiating instances of a common class is 7.78, and the average length of minimal jumping emerging patterns differentiating instances of different classes is 3.03. The experiments involved repeated mining of minimal jumping emerging patterns that occur in an instance but not occurring in small random set of several hundreds of instances.

---

[2]One may use Laplacian smoothing to change $\infty$ to very large numbers while still having higher $\mathsf{growthRate}$ values for jumping emerging patterns than for all other emerging patterns.

To get a rough idea of what the above observation means, consider this example. To differentiate residential buildings and agricultural crops one would use emerging patterns of length 1, e.g. crops grow but buildings do not. To differentiate one residential building from a collection of other residential buildings one may need to use emerging patterns of length 3 or larger. This is due to the fact that each single-attribute condition satisfied by one residential building is also satisfied by many other residential buildings; only combinations of several single-attribute conditions can indicate the uniqueness of a particular residential building.

The intrusion detection method discussed below makes use of this length difference.

### 4.2 What Emerging Patterns to Use and Their Mining

To use the observation above in an emerging pattern length based intrusion detection method, three issues need to be resolved:

(a) What kinds of emerging patterns should be used?

(b) Should we use the eager approach or lazy approach to mine the needed emerging patterns?

(c) How to compute the needed emerging patterns to increase robustness and efficiency of the intrusion detection system?

Recall that we consider one-class training based intrusion detection – we only have one training dataset $N$ representing the set of normal instances. Moreover, the number of training instances can be very large (e.g. $> 10,000$).

The discussion in Section 4.1 already hinted that we need to differentiate instances of a common class as well as differentiating instances of different classes. In particular, when predicting if a new instance is an intruder, we need to mine emerging patterns that differentiate this instance from the normal instances. In the training process, we must utilize the training data to gather needed information for use by the system to predict on new instances. Combining these, we see that we need to mine emerging patterns that differentiate an instance from a set of instances. So one possible (and our chosen) answer to Question (a) is: Use jumping emerging patterns that occur in one particular instance but not in a set of normal instances.

Because of the above choice, we need to use the lazy learning approach, where we mine patterns after we see the test instance. Moreover, in this lazy approach we only want to use statistics obtained in the training process, and we do not want to store/use all the emerging patterns mined in that process. (There are too many such patterns.)

For efficiency reasons, the emerging patterns to be used should be computed quickly to make the intrusion detection system efficient to use. It is also desirable to find ways to increase the robustness of the system. Our answer to Question (c) is to use the **BorderDiff** algorithm [9] repeatedly in a 1-vs-m fashion: one instance $t$ vs a random sample $M$ from $N$ of some suitable size $m$. This helps improve on both robustness and efficiency. By having $m$ to be a reasonably small number we get efficiency. By finding jumping emerging patters that occur in $t$ but not in $M$ for multiple $M$, the corresponding derived statistics on the patterns becomes more stable.

We note that jumping emerging patterns mined from one particular "$t$ vs $M$" may not be jumping emerging patterns for "$t$ vs $N$."

### 4.3 The OCLEP+ Training and Testing Algorithms

Given a set S of mined emerging patterns, let *minLength*(S) denote the minimal length of the patterns in S.

The training algorithm for **OCLEP+** will derive just one value: a cutoff value for the minimal lengths. For this, $k$ instances $t1, ..., tk$ are randomly drawn from $N$; each will contribute a minimal length value. For each $tj$, $r$ random samples $M_i$, each containing $m$ instances, are drawn randomly from $N - \{tj\}$. The minimal jumping emerging patterns from **BorderDiff**($tj, M_1$), ...**BorderDiff**($tj, M_r$) are collected and used to compute a minimal length $ml_{tj}$. Having $r$ invocations of **BorderDiff** on $tj$ vs $r$ different $M_i$'s helps improve the performance and robust- ness of the detection system. The list of the $k$ minimal lengths from the $k$ $tj$s are sorted to yield the cutoff value at the $p$th percentile.

---

**Algorithm 1** OCLEP+ Training

    Parameters: $k$ (size of $T$), $m$ (size of $M$), $r$ (number of $M$ per $t$),
       $p$ (percentile for cutoff value)
    Input: $N$ (dataset of normal instances)
    Output: A cutoff value $\kappa$
       *(1)* Pick a random subset $T = \{t1, ..., tk\}$ of size $k$ from $N$
       *(2)* For each $t \in T$ do
          *(3)* For each $i \in \{1, ..., r\}$ do
             *(4)* Pick a random subset $M_i$ of size $m$ from $N - \{t\}$
             (5) Let $PS_i$ be the set of minimal jumping emerging patterns
                 computed by **BorderDiff**($t, M_i$)
          (6) End
          (7) Let $ml_t = min(\{|P| \mid P \in \cup_i PS_i)$
       (8) End
       (9) Sort the list $ml_{t1}, ..., ml_{tk}$ in decreasing order
       (10) Return the cutoff value $\kappa$ at $p$-percentile of the sorted list

---

The suitable size $m$ is determined as follows: If $N$ is large, we choose $m$ in the range of [200, 800]. If $N$ is not very large, we choose $m$ to be $|N|-1$. Choosing a larger $m$ allows us to use more instances as background data to be compared against $t$, but more computational time is needed, without the benefit of additional discriminative information (compared against smaller $m$). If $m$ is roughly $N$ then the detection system may not be as robust (compared against a medium $m$), as only one set of jumping emerging patterns is available. Having $m$ in the hundreds range usually offers good speed-information tradeoff and better robustness. In experiments on the NSL-KDD data, parameters for OCLEP+ were set as follows: $k = 800$, $m = 400$, $r = 7$, and $p = 95\%$.

---
[3]OCLEP used the average length measure.

**Algorithm 2** OCLEP+ Testing
---
    Parameters: $m$ (size of $M$), $r$ (number of $M$), $\kappa$ (the cutoff value)
    Input: $N$ (dataset of normal instances), $x$ (test instance)
    Output: Is $x$ normal or intruder?
        (1) For each $i \in \{1, ..., r\}$ do
            (2) Pick a random subset $M_i$ of size $m$ from $N - \{x\}$
            (3) Let $PS_i$ be the minimal jumping emerging patterns
                computed by **BorderDiff**($x, M_i$)
        (4) End
        (5) Let $ml_x = min(\{|P| \mid P \in \cup_{i=1}^{r} PS_i\})$
        (6) If $ml_x > \kappa$ then classify $x$ as normal
        (7) Else classify $x$ as intruder
---

## 5 Experimental Evaluation of OCLEP+

In the experiments, parameters for OCLEP+ were set as follows: $k = 800$, $m = 400$, $r = 7$, and $p = 95\%$. To use pattern mining, the equi-width discretization method was used on the numerical attributes. Observe that in one-class training we do not have access to the classes.

OCLEP+ is compared against One Class SVM with linear, polynomial, and RBF kernels, and also OCLEP. The reason to choose the above three versions of SVM for evaluation is that they are very popular and also they use only one class normal data to train the classifier; we do not consider many other classifiers and techniques since they use multi-class data (both the normal and anomaly classes) to train their classifiers.

### 5.1 Details of the NSL-KDD Dataset

The NSL-KDD dataset[4] was used to evaluate OCLEP+'s performance. It is an improved version of the KDDCUP dataset (available at the UCI Repository); the modifications were performed to remove certain design deficiencies of the KDDCUP data [21]. KDDCUP was used for the Third International Knowledge Discovery and Data Mining Tools Competition, which was held in conjunction with KDD-99; the dataset includes a wide variety of intrusions that are simu-lated in a military network environment. Both KDDCUP and NSL-KDD have been widely used in evaluating intrusion detection algorithms.

---
[4]NSL-KDD: http://www.unb.ca/cic/research/datasets/nsl.html

The instances of NSL-KDD have 41 features. The NSL-KDD dataset includes a file (KDDTrain+ 20Percent) that contains both anomaly as well as normal instances. For one-class training for **OCLEP+**, the anomaly instances were removed from this file, yielding a one-class training file with 13499 normal instances. The KDDTest+ file of NSL-KDD containing 22544 instances (both normal and abnormal) was used to evaluate the classifiers.

## 5.2 Accuracy Measures

Precision, recall, F-score and accuracy are the metrics we use to compare the prediction performance of the algorithms. For any classification algorithm, the following four numbers characterize its classification performance:

- TP: The number of positive instances predicted as positives.
- FP: The number of negative instances predicted as positives.
- TN: The number of negative instances predicted as negatives.
- FN: The number of positive instances predicted as negatives.

Accuracy is defined as the proportion of the correct results:

$$Accuracy = (TP + TN)/(TP + TN + FP + FN)$$

Accuracy is not adequate for situations where the two classes are not balanced. F-score (below) is widely used for performance evaluation on imbalanced data.

F-score is the harmonic mean of both recall and precision:

$$\text{F-score} = 2 * (precision * recall)/(precision + recall)$$

where

$$Precision = TP/(TP + FP)$$
$$Recall = TP/(TP + FN)$$

## 5.3 Intrusion Detection on NSL-KDD Dataset

In general, experiments found that **OCLEP+** achieved better results than competing methods on the NSL-KDD dataset. The results are summarized in Tables 1 and 2. The tables compare performance of **OCLEP+, OCLEP,** and one-class SVM (using implementation from [5]) with three types of kernels. Table 1 gives raw counts for the confusion table, whereas Table 2 gives results on several accuracy measures.

| Method | TP | FP | TN | FN |
|---|---|---|---|---|
| OCLEP+ | 9810 | 1358 | 8353 | 3023 |
| OCLEP | 9762 | 1724 | 7987 | 3071 |
| OneClass SVM: Linear | 10615 | 4593 | 5118 | 2218 |
| OneClass SVM: Poly | 10859 | 4661 | 5050 | 1974 |
| OneClass SVM: RBF | 12825 | 9706 | 5 | 8 |

Table 1: Performance Comparison: Confusion Table

| Method | Precision | Recall | F-score | Accuracy |
|---|---|---|---|---|
| OCLEP+ | 87.84 | 76.44 | 81.75 | 80.57 |
| OCLEP | 84.99 | 76.07 | 80.28 | 78.73 |
| OneClass SVM: Linear | 69.80 | 82.72 | 75.71 | 69.79 |
| OneClass SVM: Poly | 69.97 | 84.62 | 76.60 | 70.57 |
| OneClass SVM: RBF | 56.92 | 99.94 | 72.53 | 56.91 |

Table 2: Performance Comparison on Accuracy Measures

### 5.4 Impact of Parameters

The **OCLEP+** algorithm has three parameters, namely $k$, $r$, and $m$. For a given application one should perform cross validation experiments to select the optimal parameter values. This section discusses how different values for the parameters impact the performance on the NSL-KDD dataset. As will be seen below, the optimal results are obtained when $k = 800$, $r = 7$ and $m = 400$.

The **OCLEP+** algorithm used $p$ with default value of 0.95, and the cutoff value of the minimal length statistics is 3. Any instance whose minimal length statistics is 3 or larger is classified as normal, and it is classified as anomaly otherwise.

**Impact of $k$**

We ran experiments with different values of $k$ with fixed values of $m = 400$ and $r = 7$. The results of experiment are shown in Table 3. The experimental results show that the optimum result is obtained when $k \geq 700$. We note that the minimal length of JEPs is a positive integer for each experiment. Only two minimal lengths were selected by OCLEP+ in all experiments reported in Table 3, leading to just two possible rows in the table.

| $k$  | TP   | FP   | TN   | FN   | FPR  | TPR  | Prec  | Reca  | Fscore | Accu  |
|------|------|------|------|------|------|------|-------|-------|--------|-------|
| 100  | 4301 | 565  | 9146 | 8532 | 0.06 | 0.34 | 88.39 | 33.52 | 48.60  | 59.65 |
| 200  | 4301 | 565  | 9146 | 8532 | 0.06 | 0.34 | 88.39 | 33.52 | 48.60  | 59.65 |
| 300  | 4301 | 565  | 9146 | 8532 | 0.06 | 0.34 | 88.39 | 33.52 | 48.60  | 59.65 |
| 400  | 4301 | 565  | 9146 | 8532 | 0.06 | 0.34 | 88.39 | 33.52 | 48.60  | 59.65 |
| 500  | 4301 | 565  | 9146 | 8532 | 0.06 | 0.34 | 88.39 | 33.52 | 48.60  | 59.65 |
| 600  | 4301 | 565  | 9146 | 8532 | 0.06 | 0.34 | 88.39 | 33.52 | 48.60  | 59.65 |
| 700  | 9822 | 1360 | 8351 | 3011 | 0.14 | 0.77 | 87.84 | 76.54 | 81.80  | 80.61 |
| 800  | 9822 | 1360 | 8351 | 3011 | 0.14 | 0.77 | 87.84 | 76.54 | 81.80  | 80.61 |
| 900  | 9822 | 1360 | 8351 | 3011 | 0.14 | 0.77 | 87.84 | 76.54 | 81.80  | 80.61 |
| 1000 | 9822 | 1360 | 8351 | 3011 | 0.14 | 0.77 | 87.84 | 76.54 | 81.80  | 80.61 |

Table 3: Experimental results: impact by $k$

**Impact by $r$**

We ran experiments with $r$ ranging from 1 to 10 and fixed values of $k = 800$ and $m = 400$. The results of the experiments are shown in Table 4. It can be observed that F-score value peaked when $r = 7$.

| $r$ | TP   | FP   | TN   | FN   | FPR  | TPR  | Prec  | Reca  | Fscore | Accu  |
|-----|------|------|------|------|------|------|-------|-------|--------|-------|
| 1   | 9127 | 976  | 8735 | 3706 | 0.10 | 0.71 | 90.34 | 71.12 | 79.59  | 79.23 |
| 2   | 9515 | 1136 | 8575 | 3318 | 0.12 | 0.74 | 89.33 | 74.14 | 81.03  | 80.24 |
| 3   | 9656 | 1231 | 8480 | 3177 | 0.13 | 0.75 | 88.69 | 75.24 | 81.42  | 80.45 |
| 4   | 9727 | 1279 | 8432 | 3106 | 0.13 | 0.76 | 88.38 | 75.80 | 81.61  | 80.55 |
| 5   | 9766 | 1312 | 8399 | 3067 | 0.14 | 0.76 | 88.16 | 76.10 | 81.69  | 80.58 |
| 6   | 9787 | 1348 | 8363 | 3046 | 0.14 | 0.76 | 87.89 | 76.26 | 81.67  | 80.51 |
| 7   | 9807 | 1358 | 8353 | 3026 | 0.14 | 0.76 | 87.84 | 76.42 | 81.73  | 80.55 |
| 8   | 9814 | 1374 | 8337 | 3019 | 0.14 | 0.76 | 87.72 | 76.47 | 81.71  | 80.51 |
| 9   | 9838 | 1384 | 8327 | 2995 | 0.14 | 0.77 | 87.67 | 76.66 | 81.80  | 80.58 |
| 10  | 9831 | 1408 | 8303 | 3002 | 0.14 | 0.77 | 87.47 | 76.61 | 81.68  | 80.44 |

Table 4: Experimental results: impact by $r$

**Impact by $m$**

We ran experiments with $m$ ranging from 100 to 1000 and fixed values of $k = 800$ and $r = 7$. The results of the experiments are shown in Table 5. It can be observed that increasing $m$ leads to worse recall rate but better precision. It can also be observed that the F-score is maximum at $m = 400$.

| $l$ | TP | FP | TN | FN | FPR | TPR | Prec | Reca | Fscore | Accu |
|---|---|---|---|---|---|---|---|---|---|---|
| 100 | 10388 | 3189 | 6522 | 2445 | 0.33 | 0.81 | 76.51 | 80.95 | 78.67 | 75.01 |
| 200 | 9935 | 2005 | 7706 | 2898 | 0.21 | 0.77 | 83.21 | 77.42 | 80.21 | 78.25 |
| 300 | 9846 | 1529 | 8182 | 2987 | 0.16 | 0.77 | 86.56 | 76.72 | 81.35 | 79.97 |
| 400 | 9801 | 1363 | 8348 | 3032 | 0.14 | 0.76 | 87.79 | 76.37 | 81.69 | 80.50 |
| 500 | 9758 | 1288 | 8423 | 3075 | 0.13 | 0.76 | 88.34 | 76.04 | 81.73 | 80.65 |
| 600 | 9705 | 1222 | 8489 | 3128 | 0.13 | 0.76 | 88.82 | 75.63 | 81.69 | 80.70 |
| 700 | 9621 | 1187 | 8524 | 3212 | 0.12 | 0.75 | 89.02 | 74.97 | 81.39 | 80.49 |
| 800 | 9559 | 1129 | 8582 | 3274 | 0.12 | 0.74 | 89.44 | 74.49 | 81.28 | 80.47 |
| 900 | 9468 | 1059 | 8652 | 3365 | 0.11 | 0.74 | 89.94 | 73.78 | 81.06 | 80.38 |
| 1000 | 9372 | 1025 | 8686 | 3461 | 0.11 | 0.73 | 90.14 | 73.03 | 80.69 | 80.10 |

Table 5: Experimental results: impact by $m$

## 6 Other Methods Using Emerging Patterns

We note that emerging patterns have also been called contrast patterns in the literature.

Researchers have proposed numerous emerging/contrast pattern based methods for classification [13], clustering and clustering quality evaluation [14, 18], outlier detection [6], gene ranking [19, 20], cancer analysis [17], chemoinformatics [2], and pattern aided regression/classification [11, 12]. Reference [8] is a collection on the mining and application of contrast/emerging patterns. Reference [7] gives a unified coverage of methods using emerging patterns, and a wide range of applications of those methods (which were reported in hundreds of papers); the book is written from the perspective of exploiting the power of group differences.

## 7 Concluding Remarks

This paper presented the OCLEP+ method for intrusion detection using one-class training. The method is based on the use of minimal length of jumping emerging patterns. It does not use a mathematical model, and it does not use distance metrics. Experiments showed that OCLEP+ outperformed one-class SVM with linear, polynomial, and RBF kernels.

## References

[1] Charu C Aggarwal and Philip S Yu. Outlier detection for high dimensional data. In *ACM SIGMOD Record*, volume 30, pages 37–46. ACM, 2001.

[2] Jens Auer and Jurgen Bajorath. Emerging chemical patterns: A new methodology for molecular classification and compound selection. *Journal of Chemical Information and Modeling*, 46(6):2502–2514, 2006.

[3] Kevin Beyer, Jonathan Goldstein, Raghu Ramakrishnan, and Uri Shaft. When is nearest neighbor meaningful? In *International Conference on Database Theory*, pages 217–235. Springer, 1999.

[4] Anna L Buczak and Erhan Guven. A survey of data mining and machine learning methods for cyber security intrusion detection. *IEEE Communications Surveys & Tutorials*, 18(2):1153–1176, 2016.

[5] Chih-Chung Chang and Chih-Jen Lin. Libsvm: a library for support vector machines. *ACM transactions on intelligent systems and technology (TIST)*, 2(3):27, 2011.

[6] Lijun Chen and Guozhu Dong. Masquerader detection using OCLEP: One class classification using length statistics of emerging patterns. In *Int'l Workshop on Information Processing over Evolving Networks (WINPEN)*, 2006.

[7] Guozhu Dong. *Exploiting the Power of Group Differences: Data Mining with Emerging Patterns*. Morgan & Claypool, 2019.

[8] Guozhu Dong and James Bailey, editors. *Contrast Data Mining: Concepts, Algorithms, and Applications*. Data Mining and Knowledge Discovery Series. Chapman & Hall/CRC, 2012.

[9] Guozhu Dong and Jinyan Li. Efficient mining of emerging patterns: Discovering trends and differences. In *Proc. of ACM Conf. on Knowledge Discovery and Data Mining (KDD)*, pages 43–52, 1999.

[10] Guozhu Dong and Jinyan Li. Mining border descriptions of emerging patterns from dataset pairs. *Knowl. Inf. Syst.*, 8(2):178–202, 2005.

[11] Guozhu Dong and Vahid Taslimitehrani. Pattern-aided regression modeling and prediction model analysis. *IEEE Trans. Knowl. Data Eng. (TKDE)*, 27:2452–2465, 2015.

[12] Guozhu Dong and Vahid Taslimitehrani. Pattern aided classification. In *Proceedings of SIAM International Conference on Data Mining*, pages 225–233. SIAM, 2016.

[13] Guozhu Dong, Xiuzhen Zhang, Limsoon Wong, and Jinyan Li. CAEP: Classification by aggregating emerging patterns. In *Proc. of Discovery Science*, pages 30–42, 1999.

[14] Neil Fore and Guozhu Dong. *CPC: A Contrast Pattern Based Clustering Algorithm. Chapter in Contrast Data Mining: Concepts, Algorithms and Applications, Guozhu Dong and James Bailey eds.* Chapman & Hall/CRC, 2013.

[15] Jiawei Han, Jian Pei, and Micheline Kamber. *Data mining: concepts and techniques*. Elsevier, 2011.

[16] Victoria Hodge and Jim Austin. A survey of outlier detection methodologies. *Artificial intelligence review*, 22(2):85–126, 2004.

[17] Jinyan Li and Limsoon Wong. Identifying good diagnostic gene groups from gene expression profiles using the concept of emerging patterns. *Bioinformatics*, 18(10):1406–1407, 2002.

[18] Qingbao Liu and Guozhu Dong. CPCQ: Contrast pattern based clustering quality index for categorical data. *Pattern Recognition*, 45(4):1739–1748, 2012.

[19] Shihong Mao and Guozhu Dong. Discovery of highly differentiative gene groups from microarray gene expression data using the gene club ap- proach. *Journal of Bioinformatics and Computational Biology*, 3(6):1263–1280, 2005.

[20] Shihong Mao and Guozhu Dong. *Towards Mining Optimal Emerging Patterns Amidst 1000s of Genes. Chapter in Contrast Data Mining: Concepts, Algorithms and Applications, Guozhu Dong and James Bailey eds*. Chapman & Hall/CRC (Data Mining and Knowledge Discovery Series), 2012.

[21] Mahbod Tavallaee, Ebrahim Bagheri, Wei Lu, and Ali A Ghorbani. A detailed analysis of the KDD CUP 99 data set. In *IEEE Symposium on Computational Intelligence for Security and Defense Applications (CISDA)*, pages 1–6. IEEE, 2009.